\documentclass[journal]{IEEEtran}
\usepackage{amsmath,amsfonts}
\usepackage{algorithmic}
\usepackage{algorithm}
\usepackage{array}
\usepackage[caption=false,font=normalsize,labelfont=sf,textfont=sf]{subfig}
\usepackage{textcomp}
\usepackage{stfloats}
\usepackage{url}
\usepackage{verbatim}
\usepackage{graphicx}
\usepackage{cite}
\usepackage[graphicx]{realboxes}
\usepackage{stfloats}
\usepackage{booktabs}
\usepackage{fontawesome}
\usepackage{amsmath}
\usepackage{amssymb}
\usepackage{multirow}
\usepackage[table,xcdraw]{xcolor}
\usepackage{algorithm}
\usepackage[colorlinks,linkcolor=red]{hyperref}

\usepackage[utf8]{inputenc}
\usepackage{color}

\begin{document}
%

\title{Physics Inspired Criterion \\for Pruning-Quantization Joint Learning}

\author{Weiying~Xie\href{https://orcid.org/0000-0001-8310-024X}{\includegraphics[scale=0.08]{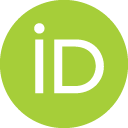}},~\IEEEmembership{Senior Member,~IEEE},
        Xiaoyi~Fan\href{https://orcid.org/0000-0001-5135-9855}{\includegraphics[scale=0.08]{./ORCIDiD.png}},
        Xin~Zhang\href{https://orcid.org/0000-0002-6455-047X}{\includegraphics[scale=0.08]{./ORCIDiD.png}},
        Yunsong~Li\href{https://orcid.org/0000-0002-0234-6270}{\includegraphics[scale=0.08]{./ORCIDiD.png}},~\IEEEmembership{Member,~IEEE} 
        \\ and Leyuan~Fang\href{https://orcid.org/0000-0003-2351-4461}{\includegraphics[scale=0.08]{./ORCIDiD.png}},~\IEEEmembership{Senior Member,~IEEE}

\thanks{This work was supported in part by the National Natural Science Foundation of China under under Grant 62121001, Grant 62322117, Grant 62371365 and Grant U22B2014; in part by Young Elite Scientist Sponsorship Program by the China Association for Science and Technology under Grant 2020QNRC001. \textit{(Corresponding author: Xiaoyi Fan)}}
\thanks{W. Xie, X. Fan, X. Zhang and Y. Li are with the State Key Laboratory of Integrated Services Networks, Xidian University, Xi’an 710071, China (e-mail: wyxie@xidian.edu.cn; xiaoyifan\_xdu@163.com; xinzhang\_xd@163.com; ysli@mail.xidian.edu.cn).} 
\thanks{L. Fang is with the College of Electrical and Information Engineering, Hunan University, Changsha 410082, China (e-mail: fangleyuan@gmail.com).}}

\markboth{IEEE TRANSACTIONS ON NEURAL NETWORKS AND LEARNING SYSTEMS, VOL. X, NO. X, X 2024}%
{Shell \MakeLowercase{\textit{et al.}}: Bare Demo of IEEEtran.cls for IEEE Journals}

\maketitle
\begin{abstract}
Pruning-quantization joint learning always facilitates the deployment of \textcolor[rgb]{0,0,0}{convolutional neural networks (CNNs)} on resource-constrained edge devices. However, most existing methods do not jointly learn a global criterion for pruning and quantization in an interpretable way. In this paper, we propose a novel physics inspired criterion for pruning-quantization joint learning (PIC-PQ), which is explored from an analogy we first draw between elasticity dynamics (ED) and model compression (MC). Specifically, derived from Hooke’s law in ED, we establish a linear relationship between the filters' importance distribution and the filter property (FP) by a learnable deformation scale in the physics inspired criterion (PIC). Furthermore, we extend PIC with a relative shift variable for a global view. To ensure feasibility and flexibility, available maximum bitwidth and penalty factor are introduced in quantization bitwidth assignment. Experiments on benchmarks of image classification demonstrate that PIC-PQ yields a good trade-off between accuracy and bit-operations (BOPs) compression ratio (\textit{e.g.,} 54.96× BOPs compression ratio in ResNet56 on CIFAR10 with 0.10\% accuracy drop and 53.24× in ResNet18 on ImageNet with 0.61\% accuracy drop). The code will be available at \url{https://github.com/fanxxxxyi/PIC-PQ}.
\end{abstract}
\begin{IEEEkeywords}

 Model compression, image classification, pruning-quantization joint learning, physics inspired criterion.

\end{IEEEkeywords}

\IEEEpeerreviewmaketitle

\section{Introduction}\label{s1}

\IEEEPARstart{S}{ignificant} advancements of CNNs have been achieved in various computer vision tasks \cite{2017Faster,li2019deep,2019DistInit,2020D2Det}, thanks to the deeper and the wider architecture of models trained on large-scale datasets. However, an issue caused by this is the requirements of computing power and memory usage, making CNNs incompatible with mobile platforms or embedded devices for real-time inference. In this context, a series of network compression techniques are proposed. Network pruning essentially derives a compact sub-network \cite{ding2019centripetal,he2019filter,lin2020hrank,chin2020towards}, and quantization converts expensive floating-point operations into hardware-friendly low-bits operations \cite{zhou2021octo,liu2020reactnet,wang2021generalizable}, both of which are beneficial for deploying CNNs on resource-constrained edge devices. 

Practical deployments often require a combination of pruning and quantization. Typically, Han \textit{et al.} \cite{han2015deep} and Louizos \textit{et al.} \cite{louizos2017bayesian} used a two-stage compression strategy, applying pruning and quantization to the model in separate sequences. Despite the resource-efficient nature of these methods in producing compressed CNNs, this disjointed learning process could potentially result in a sub-optimal solution as the two are unable to maximize their individual strengths through cooperation. Compared to two-stage compression strategy, pruning-quantization joint learning is more likely to ensure mutual promotion. For example, Ye \textit{et al.} \cite{ye2018unified} and Tung \textit{et al.} \cite{tung2018clip} practiced, but they manually set the compression ratio locally within each layer, which casts a cloud over such methods. Yang \textit{et al.} \cite{yang2020automatic} addressed the problem of automatic compression, but unstructured pruning leads to limitations in its hardware-friendliness. Wang \textit{et al.} \cite{wang2020differentiable} presented a differentiable joint pruning and quantization scheme for hardware efficiency, but neglected the concept of globally cross-layers learning. Besides, the lack of interpretability hinders the wide application of compact CNNs onto resource-constrained edge devices. In this case, it is an important yet unresolved issue to achieve both automatic and hardware-friendly pruning-quantization joint learning in an interpretable manner.

Assuming that each convolutional layer can be analogized as an elastomer, we make the first attempt to draw an analogy between the internal dynamic processes in the ED and MC field in this paper. From a macroscopic perspective, in ED, the design concept of lightweight elastomer aims to achieve the same effects as the original one while providing additional advantages in terms of cost savings. Analogously, in MC, the compressed network is expected to have the comparable level of accuracy as the baseline, but significantly reduces model size and computational complexity. From a microscopic perspective, the elastomer stretches or shortens linearly as it deforms within its elastic limit when subjected to a force. This is the well-known Hooke's law. Similarly, the filters in each layer are scattered or compactly distributed within regularization according to importance distribution after ranking in MC. The elasticity modulus (EM) is a physical quantity that captures the essential properties of elastomer. Likewise, the filter property (FP) should remain stable under different external conditions.

Motivated by this analogy, we propose a novel physics inspired criterion for pruning-quantization joint learning (PIC-PQ). Drawing on Hooke's law that the elastomers' deformation is linearly related to EM, we heuristically establish a linear relationship between the filters' importance distribution and FP by a learnable deformation scale. To be more specific, the average rank of feature maps generated by a filter consistently remains stable and effectively reflects the information richness of feature maps \cite{lin2020hrank}. Therefore, it serves as a suitable FP. While Hooke's law may possess theoretical appeal, its direct application is limited for us due to the absence of a global optimization concept. Hooke's law can only cover the importance ranking of a filter in its own layer, instead of its importance ranking among all filters. So we further introduce a relative shift variable to rank filters across different layers globally. Then, a physics inspired criterion (PIC) is completed. 

In this way, an analogy from a physical entity to deep learning increases the interpretability. Two researches \cite{li2019exploiting,zhang2022carrying} with relatively good recognition in the field of interpretability for MC point to a similar idea: \textit{Individual feature maps within and across different layers play different roles in the network. Interpreting the network, especially the feature map importance, can well guide the quantization and/or pruning of the network elements.} We are well in line with the above idea. In response, an objective function is also put forward additionally from a mathematical theory perspective to demonstrate the viability of PIC. Moreover, structural pruning, available maximum bitwidth and penalty factor ensure our method more flexible and hardware-friendly. The main contribution of this paper can be summarized as follows:

\begin{itemize}
\item We propose a novel physics inspired criterion for pruning-quantization joint learning (PIC-PQ), which is explored from an analogy we first draw between ED and MC. PIC-PQ increase the feature interpretability of MC. 

\item Specifically, derived from Hooke’s law, we establish a linear relationship between the filters' importance distribution and FP by a learnable deformation scale in PIC. We further extend PIC with a relative shift variable for a global view. Additionally, an objective function proves the viability of PIC in terms of mathematical theory. Besides, available maximum bitwidth and penalty factor in quantization bitwidth assignment ensure the feasibility and flexibility.

\item Sufficient experiments are conducted to show the effectiveness of PIC-PQ (\textit{e.g.,} 54.96× BOPs compression ratio in ResNet56 on CIFAR10 with 0.10\% accuracy drop and 53.24× BOPs compression ratio in ResNet18 on ImageNet with 0.61\% accuracy drop).
\end{itemize}

The rest of this paper is organized as follows. Section \ref{s2} presents a brief review of related works. The proposed method is then described in detail in Section \ref{s3}. Section \ref{s4} reports the experimental results as well as comparisons with \textcolor[rgb]{0,0,0}{state-of-the-art methods (SOTAs)} on benchmarks. Section \ref{s5} discusses a range of different experiment setups. In Section \ref{s6}, we conclude the proposed method.

\section{Related Work}\label{s2}

\subsection{Pruning-Quantization Joint Learning}

\textcolor[rgb]{0,0,0}{Currently, some efforts have committed to giving a solution to this problem.} Ye \textit{et al.}  \cite{ye2018unified} and Tung \textit{et al.} \cite{tung2018clip} relied on setting hyperparameters to compress layers with the desired compression rate. Yang \textit{et al.} \cite{yang2020automatic} decoupled constrained optimization by alternating directional multiplication method (ADMM). APQ \cite{wang2020apq} proposed a method for efficient inference on resource-constrained hardware and designed a promising quantization-aware accuracy predictor. DJPQ \cite{wang2020differentiable} incorporated structured pruning based on variation information bottlenecks and hybrid accuracy quantization into a single differentiable loss function. However, these methods rarely investigate the interpretability of pruning-quantization joint learning. Recently, the interpretation theory of CNNs has demonstrated its effectiveness and superior promise by achieving SOTAs' results on various computer vision tasks \cite{li2019exploiting,zhang2022carrying}. Therefore, we further explore the feature interpretability.

\subsection{Bridging CNNs with Knowledge of Mathematics or Physics}

This may be a good breakthrough if we want to unravel the mystery behind the ``black box” of CNNs so that research results from both fields can be cross-referenced. Drawing on discrete dynamical systems, the similarity between ResNet and the discretization of ordinary differential equations (ODEs) was explored \cite{he2019ode}. Following, knowledge from fluid dynamics (FD) and heat transfer was borrowed to interpret CNNs for image super-resolution \cite{zhang2022fluid,zhang2022heat}. Understanding model compression in a mathematical or physical sense remains an open and worthwhile task, which has significant implications for the construction of a credible and interpretable artificial intelligence (AI) system. As a prominent branch of solid mechanics, ED is the study of the deformation and internal forces that occur in elastic objects under the action of external forces and other external factors. We open up a new research way for pruning-quantization joint learning.

\section{Method}\label{s3}

\subsection{The Analogy between ED and MC}\label{s3.1}

\begin{figure}[ht]
\centering
\includegraphics[scale = 0.35]{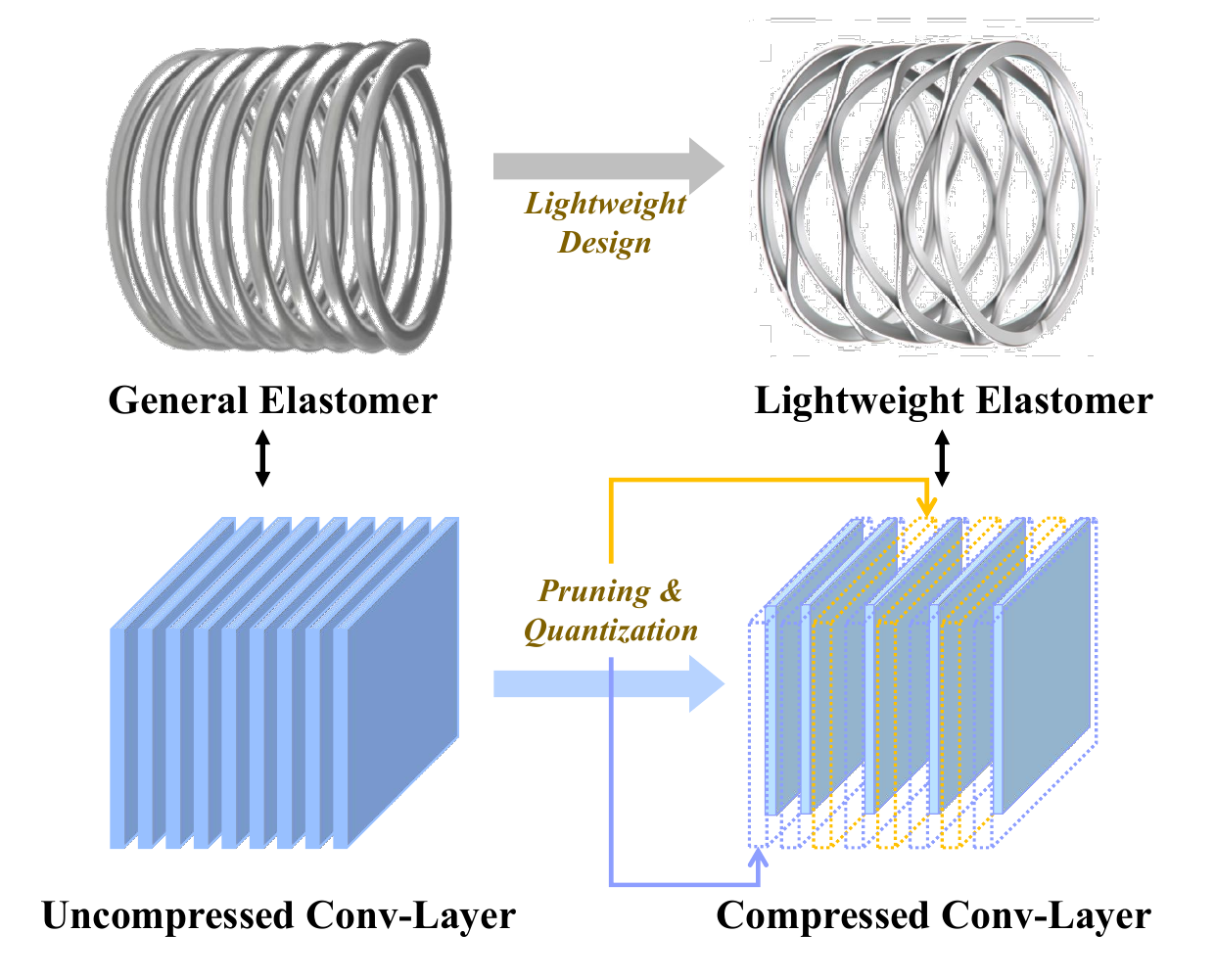}
\vskip -0.15in
\caption{The analogy between the optimization in ED and MC.}
\label{p1}
\vskip -0.15in
\end{figure}

This study is inspired by the analogy between the ED and MC. Specifically, this similarity is reflected in two aspects. From a macroscopic perspective, the lightweight elastomer aims to endure the same force and load specifications as conventional one but with a resource-friendly space. Similarly, the compressed CNN is expected to have the comparable level of accuracy as the baseline, but with significantly reduced model size and computational complexity. From a microscopic perspective, the FP is stable, just as the EM is fixed. The regularization of the compact CNN governs the filters' importance distribution, precisely like certain boundary conditions applied to the elastomers. Moreover, after regularization, the filters of each layer show a scattered or compact distribution depending on the importance ranking, like an elastomer stretching or shortening linearly within the elastic limit after external force is applied. The analogy between the optimization in ED and MC is shown in Figure \ref{p1}.

In ED, the need to reduce weight and space for engineering applications has sparked two innovative ideas for elastomer optimization. On the one hand, materials with low EM should be removed as far as possible within cost constraints and replaced by more efficient and stable materials with higher EM. On the other hand, special design can be used to save space and costs. The thickness and width of the material, or the wave number of the lightweight elastomer can be varied to meet the requirements of different loads. Similarly, on the one hand, the MC can use network pruning which only keeps those filters that are relatively important, and those who have small FP should be deleted within each layer. On the other hand, quantization converts expensive floating point operations into hardware-friendly low-bit operations. The quantization bitwidth of each layer can also be adjusted by designed parameters. Motivated by the consistency in these two fields, useful ideas can be borrowed from ED to MC. 

\subsection{Physics Inspired Criterion for Ranking Filters Globally}\label{s3.2}

\begin{figure*}[ht]
\centering
\includegraphics[scale=0.9]{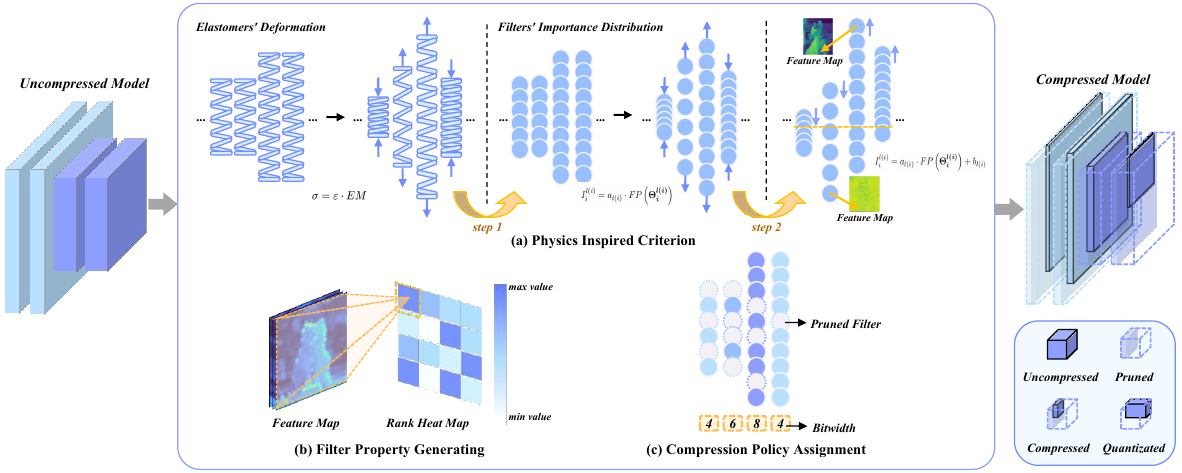}
\caption{Overview of PIC-PQ framework.  (a) Step 1 first shows a visual analogy between elastomers' deformation in ED and filters' importance distribution in MC. And it establishes a linear relationship that the filters' importance distribution is linearly related to FP. In step 2, a relative shift variable is further introduced to rank filters cross-layers globally. (b) Then, the rank of the feature map from each filter is generated for obtaining the global importance ranking with the optimal $\boldsymbol{a-b}$ pairs searched before. (c) Finally, compression policy is assigned automatically.  }
\label{p2}
\vskip -0.15in
\end{figure*}

As step 1 in Figure \ref{p2} (a) shows, Hooke's law is expressed as the elasticity of an elastomer deformed by linear stretching or shortening within the elastic limit following a force:

\begin{equation}
\sigma = \varepsilon\cdot{EM},\label{eq1}
\end{equation}
where $\sigma$ indicates the force that stretches or shortens the elastomer, $\varepsilon$ denotes the deformation by stretching or shortening, and $EM$ is a constant factor called the elasticity modulus. If we consider each convolutional layer as an elastomer, we can liken the FP to the EM, and the magnitude of the learnable scale can be viewed as the deformation. Thus, the $i$-th filter's importance distribution $I_{i}^{l(i)}$ depends on the product of a learnable deformation scale and the FP:

\begin{equation}
I_{i}^{l(i)}=a_{l(i)}\cdot FP\left(\boldsymbol{\Theta_{i}^{l(i)}} \right),\label{eq2}
\end{equation}
where $l(i)$ is the layer index for the $i$-th filter, $a_{l(i)}$ is the learnable deformation scale, $\boldsymbol{\Theta_{i}^{l(i)}}$ represents the $i$-th filter and $FP\left(\boldsymbol{\Theta_{i}^{l(i)}} \right)$ is the FP of the $i$-th filter. Although Hooke's law has given us much inspiration, it cannot be fully replicated. Thus, two issues we need to address are: (1) Hooke's law articulates an individual concept, meaning that it can only cover the importance ranking of a filter in its layer, which requires us to extend it to the global level. (2) The deformation of the elastomer is hand-crafted rather than automatic, whereas we need to further explore end-to-end automatic compression. For the first issue, a relative shift variable $b_{l(i)}$ is further introduced to rank filters cross-layers globally as shown in step 2 in Figure \ref{p2} (a):

\begin{equation}
I_{i}^{l(i)}=a_{l(i)}\cdot FP\left(\boldsymbol{\Theta_{i}^{l(i)}} \right)+b_{l(i)}.\label{eq3}
\end{equation}

\textcolor[rgb]{0,0,0}{In Figure \ref{p2} (a), we can see that the filter with the highest global importance has the most information richness, and conversely, the filter with the lowest global importance has the least information richness. This confirms the feature's interpretability.}

EM is an intrinsic property of the elastomer that does not vary with force. It reveals the need to find an information metric characterizing FP, also an intrinsic property of the filters which does not change with external conditions. It has been observed in HRank \cite{lin2020hrank} that the rank generated by a filter is robust to the input images. Therefore, a small batch of input images can be used to accurately estimate the expectation of the feature map rank as illustrated in Figure \ref{p2} (b). \textcolor[rgb]{0,0,0}{Assuming that $P(D)$ means the distribution of training dataset, while $D$ to denote the input images. The rank of feature maps is not only a valid information measure but also a stable representation across distribution $P(D)$.} The underlying principle is that feature mapping is an intermediate step that can reflect both filters' attributes and input images. Even within the same layer, different filters and their feature maps play different roles in the network \cite{li2019exploiting, zhang2022carrying}. In addition, they capture how the input images are transformed in each layer and, finally, to the predicted labels. We define the FP as:
\begin{equation}
FP\left(\boldsymbol{\Theta_{i}^{l(i)}} \right)=\mathrm{R}\left(\mathbf{o}_{i}^{l(i)}(D)\right),\label{eq4}
\end{equation}
where $\mathbf{o}_{i}^{l(i)}(D)$ is the feature map of the $i$-th filter in the $l(i)$-th layer. \textcolor[rgb]{0,0,0}{$R(\cdot)$ is the expectation of the feature maps' rank for input images $D$.} 

When inputting the $d$-th image in \textcolor[rgb]{0,0,0}{sampled} input images $D$, the singular value decomposition (SVD) of $\mathbf{o}_{i}^{l(i)}(D^d)$ is:
\begin{gather}
\begin{aligned}
&\mathbf{o}_{i}^{l(i)}(D^d) = \sum_{j=1}^{\left ( {r_{i}^{l(i)}}\right ) ^d} \delta_ju_jv_j^{T}\\
&=\sum_{j=1}^{\left ( {(r') _{i}^{l(i)}} \right )^d}  \delta_ju_jv_j^{T}+\sum_{j={\left ( {(r') _{i}^{l(i)}} \right )^d} +1}^{\left ( {r_{i}^{l(i)}}\right ) ^d} \delta_ju_jv_j^{T},\label{eq5}
\end{aligned}
\end{gather}
where $\delta_j$, $u_j$ and $v_j$ are left singular vectors, the previous $j$ singular values and right singular vectors of $\mathbf{o}_{i}^{l(i)}(D^d)$, respectively. It can be seen that the feature map with rank ${r_{i}^{l(i)}}$ can be decomposed into a low-rank feature map with rank ${(r') _{i}^{l(i)}}$ and some additional information. Thus, the high-rank feature map contains more information than the low-rank feature map. \textcolor[rgb]{0,0,0}{Eq. (\ref{eq5}) is used to show that the expectation of rank can be a reliable measure of information richness.} $\mathrm{R}\left(\mathbf{o}_{i}^{l(i)}(D)\right)$ can be calculated as:

\begin{equation}
\mathrm{R}\left(\mathbf{o}_{i}^{l(i)}(D)\right)=\frac{1}{\left | D \right |}  \sum_{d}^{\left | D \right |} {\left ( {r_{i}^{l(i)}}\right ) ^d},\label{eq6}
\end{equation}
where $\left | D \right |$ is the number of input images.

\subsection{The Derivation from a Mathematical Theory Perspective}\label{s3.3}

The PIC above is based on the analogy between the ED and MC and plus a further design. In the following, we would like to derive this method from a mathematical theory perspective additionally. We want the compressed model after a series of optimization designs not only result in significant resource savings but also maintain lower accuracy drop compared to the original model. This is in line with the original intention of the MC, \textit{i.e.}, we want the difference in the loss function of the CNN before and after compression to be as small as possible. Therefore, we give the following optimization problem:
\vskip -0.15in
\begin{gather}
\begin{aligned}
\min_{\mathbf{\mathcal{P}(m)}} \mathbb{L}\left(\boldsymbol{\Theta} \odot \mathbf{\mathcal{P}(m)}-\eta \sum_{j=1}^{\tau} \Delta \boldsymbol{w}^{(j)} \odot \mathbf{\mathcal{P}(m)}\right)-\mathbb{L}(\boldsymbol{\Theta}),
\\\text { s.t. } C(\mathbf{\mathcal{P}(m)}) \leq S_{budget},\label{eq7}
\end{aligned}
\end{gather}
\textcolor[rgb]{0,0,0}{where $\mathbf{\mathcal{P}(m)}$ denotes the pruning-quantization joint learning strategy, which can be a two-dimensional matrix composed of a function, with the row index representing the number of convolutional layers and the column index representing the filter index (with 0 in the blank places since the number of filters is not the same in every layer of a network). $\mathbf{\mathcal{P}(\cdot)}$ represents the quantization operation based on the allocated bitwidth.} The binary variable $\mathbf{m}$ denotes the filter mask, and if the importance ranking of a filter is below the pruning threshold \textcolor[rgb]{0,0,0}{(calculated from the resource constraints and the filter global ranking), $\mathbf{m}_{i}^{l(i)}=0$, $\mathbf{\mathcal{P}(m)}=0$, otherwise $\mathbf{m}_{i}^{l(i)}=1$, quantize the corresponding filter.} The Hadamard product $\odot$ represents the multiplication of elements in corresponding positions, $\eta$ and $\tau$ denote the learning rate and the number of gradient steps respectively, and $\Delta \boldsymbol{w}^{(j)}$ denotes the gradient for the filter weights computed at step $j$. In terms of constraints, $C(\cdot)$ is the modeling function for computational complexity, and $S_{budget}$ is the required computational complexity constraint. 

\textcolor[rgb]{0,0,0}{The form of the loss function $\mathbb{L}(\Theta)$ is:}
\vskip -0.15in
\begin{equation}
\mathbb{L}(\Theta)=\frac{1}{|D|} \sum_{(x, y) \in D} \sum_{cls=1}^{N} y_{\mathrm{cls}} \log \left(\mathrm{f}(x\mid \Theta)_{cls}\right),\label{eq8}
\end{equation}
\textcolor[rgb]{0,0,0}{where $x$ and $y$ are the input and label, respectively. $D$ is the training data, $cls$ is the class index, and $N$ is the number of classes. $\mathrm{f}$ denotes the CNN model, so  $\mathrm{f}(x\mid \Theta)_{cls}$ means the probability that observation sample $x$ belongs to class $cls$.}

In mathematical problems, the Lipschitz continuity condition is often used to approximate the optimization problem of complex functions into a quadratic programming problem. The Lipschitz continuity condition is defined as:

\textit{For a function $f:U\subseteq \mathbb{R}^m\to \mathbb{R}^n$, if there exists a constant $\Omega\ge 0$ such that:}
\vskip -0.15in
\begin{equation}
\left \| f(y)-f(x) \right \| \le \Omega\left \| y-x \right \| ,\forall x,y\in  U,\label{eq9}
\end{equation}
\textit{where $f$ satisfies the Lipschitz continuous, and the constant $\Omega$ is called the Lipschitz constant for $f$.}
\vskip -0.15in
\begin{figure}[h]
\centering
\includegraphics[scale=0.25]{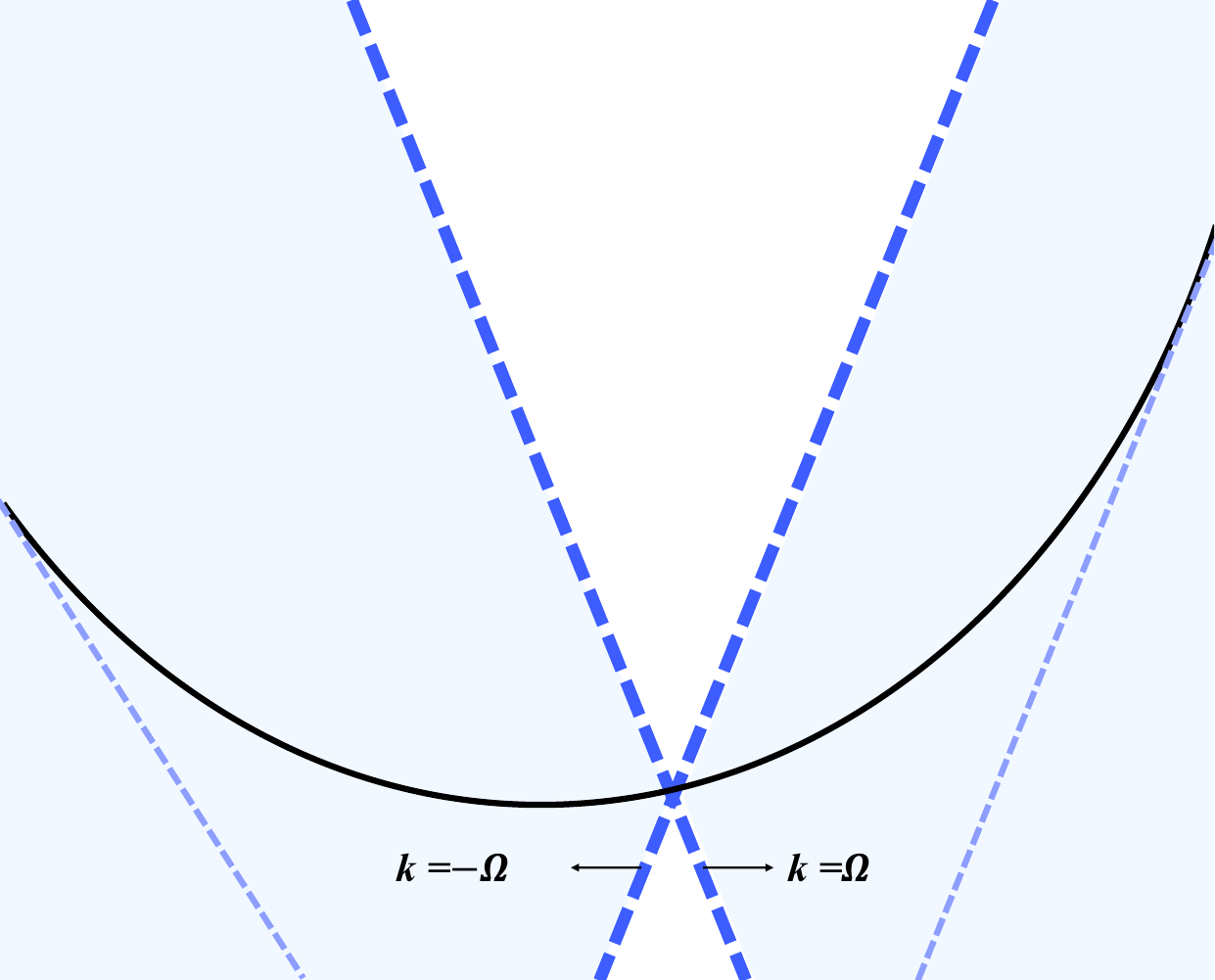}
\caption{\textcolor[rgb]{0,0,0}{Lipshitz bound on the function. $k$ denotes the slope.}}
\label{p3}
\vskip -0.15in
\end{figure}

\textcolor[rgb]{0,0,0}{As shown in the Figure \ref{p3}, $k$ denotes the slope. the Lipschitz continuous, which requires that the slopes of any two points on the curve joining the image of a function be uniformly bounded, is that any of the slopes are less than the same constant, and this constant is the Lipschitz constant. From a local point of view: we can take two sufficiently close to the point, if this time the limit of the slope exists, the limit of this slope is the derivative of this point. That is, if the function is conductible and Lipschitz continuous, then the derivative is bounded. Conversely, if the function is conductible and the derivative is bounded, it can be introduced that the function is Lipschitz continuous. So we can determine an upper bound on the difference in loss between the baseline and compressed model by Lipschitz continuous.}

The loss function $\mathbb{L}(\Theta)$ is $\Omega_l$-Lipschitz continuous for the $l$-th layer of CNN, Eq. (\ref{eq7}) can be deflated to find an upper bound:
\vskip -0.15in
\begin{gather}
\begin{aligned}
& \mathbb{L}\left(\boldsymbol{\Theta} \odot \mathbf{\mathcal{P}(m)}-\eta \sum_{j=1}^{\tau} \Delta \boldsymbol{w}^{(j)} \odot \mathbf{\mathcal{P}(m)}\right)-\mathbb{L}(\boldsymbol{\Theta}) \\\leq & \mathbb{L}(\boldsymbol{\Theta} \odot \mathbf{\mathcal{P}(m)})+\sum_{i=1}^{I} \Omega_{l(i)} \eta\left\|\sum_{j=1}^{\tau} \Delta \boldsymbol{w}_{i}^{(j)} \odot \mathbf{\mathcal{P}(m_{i}^{l(i)})}\right\|-\mathbb{L}(\boldsymbol{\Theta}) \\
\leq & \sum_{i=1}^{I} \Omega_{l(i)}\left\|\boldsymbol{\Theta_{i}^{l(i)}}\right\| (1-\mathbf{\mathcal{P}(m_{i}^{l(i)})})+\sum_{i=1}^{I} \Omega_{l(i)}^{2} \eta \tau \odot \mathbf{\mathcal{P}(m_{i}^{l(i)})}\\
=& \sum_{i=1}^{I}\left(\Omega_{l(i)}\left\|\boldsymbol{\Theta_{i}^{l(i)}}\right\|-\Omega_{l(i)}^{2} \eta \tau\right) \mathbf{\mathcal{Q} (m_{i}^{l(i)})}+\Omega_{l(i)}^{2} \eta \tau,\label{eq10}
\end{aligned}
\end{gather}
where $l(i)$ is the layer index of the $i$-th filter, $I$ indicates the total number of filters in the $l(i)$-th layer, $\mathbf{\mathcal{Q}(m)} =1-\mathbf{\mathcal{P}(m)}$, $||\cdot||$ denotes the arbitrary regularization tool.

The left side of the constraint in Eq. (\ref{eq7}) can be written in the following form:
\vskip -0.15in
\begin{gather}
\begin{aligned}
&C({\mathbf{\mathcal{P}(m)}})=\sum_{i}^{I}c_{l(i)} \mathbf{\mathcal{P}(m_{i}^{l(i)})}\left \| \bigtriangleup  \right \| _{0},c_{l(i)} \geq 0,\\
&\bigtriangleup=\left \{ [\boldsymbol{\Theta} \odot \mathbf{\mathcal{P}(m)}]:[\boldsymbol{\Theta}_j \odot\mathbf{\mathcal{P}(m_j)}] \forall j \in \Phi (l(i)) \right \},\label{eq11}
\end{aligned}
\end{gather}
where $\Phi (l(i))$ returns the set of filter indexes before layer $l(i)$ and $c_{l(i)}$ is a constant associated with the layer. Let $C_{l(i)}$ be the computational complexity count of layer $l(i)$ where filter $i$ resides, which depends linearly on the number of filters in the previous layer:

\begin{equation}
C_{l(i)}=c_{l(i)} \left \| \bigtriangleup  \right \| _{0}.\label{eq12}
\end{equation}

Then Eq. (\ref{eq11}) can be abbreviated as:
\vskip -0.15in
\begin{gather}
\begin{aligned}
&C({\mathbf{\mathcal{P}(m)}})=C(\mathbf{1}-{\mathbf{\mathcal{Q}(m)}})\\
=&\sum_{i}^{I} C_{l(i)}{\mathbf{\mathcal{P}(m_{i}^{l(i)})}}=\sum_{i}^{I} C_{l(i)}\left(1-{\mathbf{\mathcal{Q}(m_{i}^{l(i)})}}\right).\label{eq13}
\end{aligned}
\end{gather}

Let $\hat{C}_{l(i)}$ to denote the computational complexity count at layer $l(i)$ of the baseline network, and it follows that $C_{l(i)} \leq \hat{C}_{l(i)}$. Thus, there is the following form:
\vskip -0.15in
\begin{gather}
\begin{aligned}
&C({\mathbf{\mathcal{P}(m)}})=C(\mathbf{1}-\mathbf{\mathcal{Q} })\\
&=\sum_{i}^{I} C_{l(i)}\left(1-\mathcal{Q}_{i}\right) \leq \sum_{i}^{I} \hat{C}_{l(i)}\left(1-\mathcal{Q}_{i}\right).\label{eq14}
\end{aligned}
\end{gather}

Further, Eq. (\ref{eq7}) can be transformed into a Lagrange form by minimizing its upper bound:
\vskip -0.15in
\begin{equation}
\min _{\mathbf{\mathcal{Q}(m)}} \sum_{i=1}^{I}\left(\Omega_{l(i)}\left\|\boldsymbol{\Theta_{i}^{l(i)}}\right\| +\eta \tau \Omega_{l(i)}^{2}-\lambda \hat{C}_{l(i)}\right) \mathbf{\mathcal{Q}(m_{i}^{l(i)})},
\label{eq15}
\end{equation}
\textcolor[rgb]{0,0,0}{where $\lambda$ denotes the Lagrange multiplier. Let $a_{l(i)} = \Omega_{l(i)}$ and $b_{l(i)} = \eta \tau \Omega_{l(i)}^{2}-\lambda \hat{C}_{l(i)}$. $\left\|\boldsymbol{\Theta_{i}^{l(i)}}\right\|$ is approximated by $\mathrm{R}\left(\mathbf{o}_{i}^{l(i)}(D)\right)$. Therefore, the optimization problem above can be simplified as:}

\begin{equation}
\min _{\mathbf{\mathcal{Q}(m)}} \sum_{i=1}^{I}\left(a_{l(i)}\mathrm{R}\left(\mathbf{o}_{i}^{l(i)}(D)\right) +b_{l(i)}\right)\mathbf{\mathcal{Q}(m_{i}^{l(i)})}.\label{eq16}
\end{equation}

That is, the process of globally cross-layers learning filter ranking can be seen as learning to estimate $\boldsymbol{a}$ and $\boldsymbol{b}$ and solving for the optimal solution to Eq. (\ref{eq16}) to produce a better solution to the original objective. This is the same as Eq. (\ref{eq3}) that we give in imitation of ED theory. \textcolor[rgb]{0,0,0}{The analogy between} the ED and MC is a prerequisite for our formulation of the optimization objectives, and the subsequent mathematical derivation provides the theoretical support for Eq. (\ref{eq3}). This forms a closed loop perfectly.

\subsection{Quantization Bitwidth Automatically Assignment}\label{s3.4}

After rank generating, PIC-PQ obtains the global importance ranking with the optimal $\boldsymbol{a-b}$ pairs learned. As Figure \ref{p2} (c) shows, finally, compression policy is assigned automatically to ensure meeting the constraints. As the filters' importance knowledge is encoded in the $\boldsymbol{a-b}$ pairs, this process can be done efficiently without the need for training data. More importantly, there is no subjective human intervention or empirical parameter settings. Furthermore, in order to reduce the search space while ensuring an efficient hardware implementation, only different bitwidths are used between layers, but within layers, they are kept uniform.

Quantization bitwidth depends on the role of the quantized object in the overall model. For example, for over-parameterized filters, a lower quantization bitwidth can be chosen. At the same time, we also need to consider how to set the quantization bitwidth in terms of hardware resources. Based on the global importance ranking, we can determine the layer sparsity $S_l$ of $l$-th layer:
\vskip -0.1in
\begin{equation}
S_l=\frac{\left\|\mathbf{\Theta}_l\odot \mathbf{M} _l\right\|_{1}}{\left\|\mathbf{\Theta}_l\right\|_{1}},\label{eq17}
\end{equation}
where $\mathbf{\Theta}_l$ denotes the vector composed of filters in layer $l$, $\mathbf{M} _l$ denotes the sparse mask for layer $l$, which is composed of the filter mask $\mathbf{m}_{i}^{l(i)}$. 

Then the quantization bitwidths of the weights and activation functions can be calculated according to:
\vskip -0.1in
\begin{equation}
n_{W_l}=\lceil N_{W_l}-\frac{p}{S_l} \rceil, 
n_{A_l}=\lceil N_{A_l}-\frac{p}{S_l} \rceil,\label{eq18}
\end{equation}
where $n_{W_l}$ and $N_{W_l}$ respectively correspond to the bitwidth for weight quantization and the available maximum bitwidth for $l$-th layer. Similarly, $n_{A_l}$ and $N_{A_l}$ do with quantization of the activation function. Penalty factor $p$ is a constant that can be adjusted according to the hardware.

For the $n_{W_l}$-bit weight and the $n_{A_l}$-bit activation output, the quantization function is:
\vskip -0.1in
\begin{equation}
{W_l}^{q}=\frac{\operatorname{round}\left[\frac{\tanh \left({W_l}\right) {2^{n_{W_l}-1}}}{\max \left(\left|{W_l}\right|\right)}\right]}{2^{n_{W_l}-1}}.\label{eq19}
\end{equation}

\begin{equation}
{A_l}^{q}=\frac{\operatorname { round }\left[ \operatorname { clamp }\left(A_l, 0,1\right) \bullet 2^{n_{A_l}}\right]}{2^{n_{A_l}}}.\label{eq20}
\end{equation}

We believe that the different input complexity of the layers of the network implies different precision requirements for the neurons. Neurons in shallower layers should be more sensitive to quantization \cite{chu2021mixed}. Since the feature distributions overlap each other, a limited number of neurons will not be able to distinguish between samples and extract meaningful intermediate representations without the right precision. Once high-level features are explicitly obtained, later layers become more robust to quantization errors. 

\begin{figure*}[ht]
\centering 
\vskip -0.15in
\includegraphics[scale = 0.5]{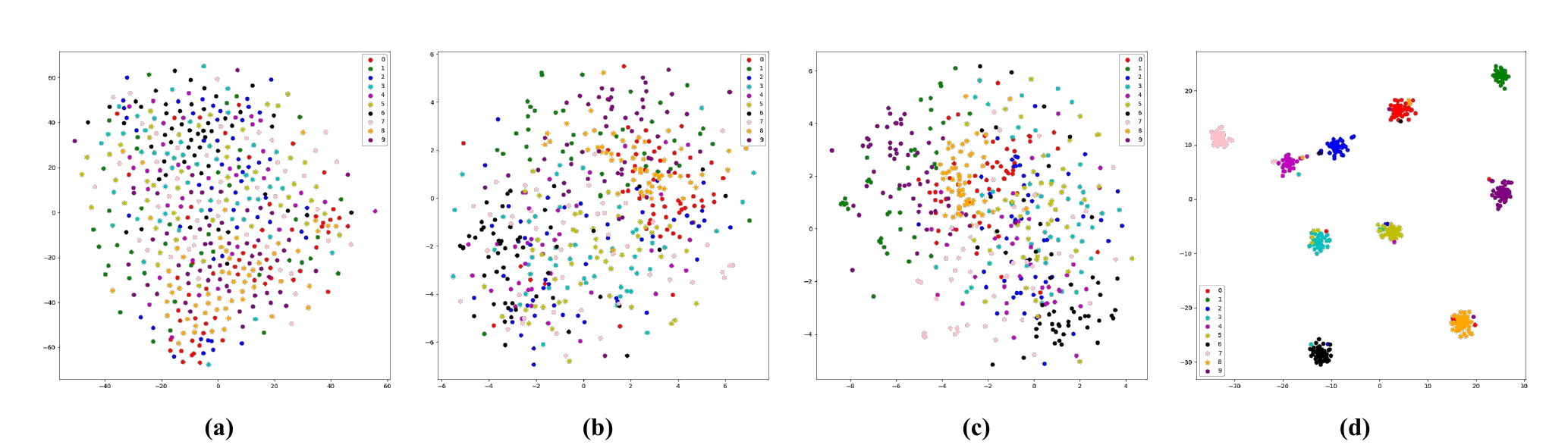}
\caption{Two-dimensional feature representation of the inner layers of the ResNet56 network trained on CIFAR10. (a) Conv1 (first layer before Block 1) (b) Conv19 (last layer of Block 1) (c) Conv37 (last layer of Block 2) (d) Conv55 (last layer of Block 3).}
\label{p4}
\vskip -0.15in
\end{figure*}

To explore further, we trained a full-precision ResNet56 on CIFAR10. Drawing on the analytical approach from Chu \textit{et al.} \cite{chu2021mixed}, 50 samples from each category are randomly selected and feed into the model at the end of training. T-SNE \cite{van2008visualizing} is used to extract the feature representation of the inner layer and transform it into two-dimensional features. As shown in Figure \ref{p4}, in the initial layer, there is significant confounding between the distribution of low-level semantic features of different categories. It is difficult to determine labels directly from the underlying features. Many fine neurons are required to distinguish between overlapping distributions. As the network propagates forward, features of the same class are gradually aggregated. As shown in Figure \ref{p4}(d), the high-level semantic features are more robust, with clear boundaries between clusters of distributions of different categories at a deeper level. Based on this observation, we tentatively corroborate the above opinion, and also consider that the shallow neurons should be more sensitive to quantization and the deeper layers more robust to quantization errors. Therefore, the available maximum bitwidth reduces as the layer deepens.

\subsection{Obtain the Compressed Model}\label{s3.5}

\begin{algorithm}[h]
   \caption{Algorithm Description of PIC-PQ.}
   \label{Algorithm Description of PIC-PQ}
    \begin{algorithmic}[1]
        \STATE{\bfseries Input:} Randomly \textcolor[rgb]{0,0,0}{sampled} images $D_{sample}$, baseline model $\boldsymbol{\Theta}$, constraint $S_{budget}$, random walk size $W$, total search iterations $E$, sample size $S$, mutation ratio $u$, population size $P$, fine-tuning iterations $\boldsymbol{\tau}$.
        \STATE Initialize $\boldsymbol{\Theta}$;
        \STATE \textcolor[rgb]{0,0,0}{Use $D_{sample}$ to generate $\textbf{R}$ of each filter's feature maps;}
        \STATE Initialize \textit{pool} to a size of $P$;
        \FOR{$e=1$ {\bfseries to} $E$}
        \STATE \textit{$\textbf{a}$}, \textit{$\textbf{b}$} = \textbf{Regularized EA} ($W$, $S$, $u$,\textit{ Pool});
        \STATE $\hat{\boldsymbol{\Theta}}^\ast$ =\textbf{ Compressing and fine-tuning} ($\boldsymbol{\Theta}$, \textit{$\textbf{a}$},\textit{$\textbf{b}$}, $\boldsymbol{\tau}$, $S_{budget}$, $\textbf{R}$);
        \STATE \textit{Score} = $Acc_{val}(\hat{\boldsymbol{\Theta}}^\ast)$;
        \STATE \textit{pool}.replaceOldestWith(\textit{$\textbf{a}$}, \textit{$\textbf{b}$}, \textit{Score});
        \ENDFOR
        \STATE Rank \textbf{filters' importance} according to Eq. (\ref{eq3});
        \STATE Choose filters to prune, and assign bitwidth $n_{W_l}, n_{A_l}$ for layers according to Eq. (\ref{eq18});
        \STATE Compress and fine-tune the model;
        \STATE{\bfseries Output:} The best compressed model $\boldsymbol{\Theta^\ast}$ with highest $Acc_{val}(\boldsymbol{\Theta^\ast})$.
    \end{algorithmic}
\end{algorithm}

The filters' importance distribution is updated by searching $\boldsymbol{a-b}$ pairs of layers. In the initial state, the distribution is unstressed and unmoving, meaning that $a_{l(i)}$ and $b_{l(i)}$ are 1 and 0 respectively. \textcolor[rgb]{0,0,0}{Compressed subnetworks of different computational complexity will be obtained by learning $\boldsymbol{a-b}$ pairs under different constraints.} Meaning as in Algorithm \ref{Algorithm Description of PIC-PQ}, we use the regularized evolutionary algorithm (Regularized EA) proposed in \cite{real2019regularized} to learn $\boldsymbol{a}$ and $\boldsymbol{b}$. \textcolor[rgb]{0,0,0}{We first generate a \textit{pool} of candidates ($\boldsymbol{a-b}$ pairs) and record the \textit{fitness score} for each candidate, and then repeat the following steps during the total search iterations: (1) Sample a subset to a size of S from the candidates, and identify the fittest candidate. Then generate a new candidate by mutating the fittest candidate of mutation ratio $u$. (2) each $\boldsymbol{a-b}$ pair generated from Regularized EA equates to a compressed architecture, and we fine-tune the sub-network employing the $\boldsymbol{\tau}$ gradient step.} \textcolor[rgb]{0,0,0}{Note that we use $\boldsymbol{\tau}$ to approximate $\boldsymbol{\hat{\tau}}$ (a fully fine-tuned step) and we empirically find that $\boldsymbol{\tau}=200$ works well. (3) Then evaluate validation accuracy as \textit{fitness score} of the new candidate. (4) Replace the oldest candidate in the \textit{pool} with the generated $\boldsymbol{a-b}$ pair and its \textit{fitness score}.} 

\section{Experiments}\label{s4}

In this section, we conduct extensive experiments to demonstrate the effectiveness of our method. We first present the experiment setting in Section \ref{s4.1}, and then we show the experiments of PIC-PQ on image classification benchmarks and compare them with SOTAs in Section \ref{s4.2}. 

\begin{table*}[ht]
\caption{Comparison of Compressing ResNet56 and VGG16 on CIFAR10}
\label{table-CIFAR10}
\begin{center}
\scalebox{1.15}{
\begin{tabular}{ccccc|ccccc}
\toprule
\multicolumn{5}{c|}{\textbf{ResNet56}} & \multicolumn{5}{c}{\textbf{VGG16}} \\ \cline{1-10}
\textbf{Method}  & \textbf{P} & \textbf{Q}   & \textbf{Top-1(\%)↓} & \textbf{Comp. Ratio} &\textbf{Method}  & \textbf{P} & \textbf{Q}   & \textbf{Top-1(\%)↓} & \textbf{Comp. Ratio}\\
\midrule

HRank \cite{lin2020hrank}              &$\surd$     &$\times$     & -0.26          & 1.41             &LRMF\cite{zhang2021filter}                      &$\surd$     &$\times$     & -0.12  & 1.56 \\    
\textbf{PIC-P}                         &$\surd$     &$\times$     & \textbf{-0.93} & \textbf{1.43}    &Zhao \textit{et al.} \cite{zhao2019variational} &$\surd$     &$\times$     & 0.78  & 1.64\\
REAF\cite{zhang2023reaf} &$\surd$     &$\times$     & -0.62  & 1.61                                   &GAL-0.05 \cite{lin2019towards}                  &$\surd$     &$\times$     & 1.93  & 1.66\\
CCM-LRR\cite{wang2023filter} &$\surd$     &$\times$     & -0.25  & 1.62                               &SSS \cite{huang2018data}                        &$\surd$     &$\times$     & 0.94  & 1.71 \\
HBFP\cite{basha2024deep} &$\surd$     &$\times$     & 0.84  & 1.78                                    &GAL-0.1 \cite{lin2019towards}                   &$\surd$     &$\times$     & 3.23  & 1.82 \\
REAF\cite{zhang2023reaf} &$\surd$     &$\times$     & -0.08  & 2.00                                   &CCM-LRR\cite{wang2023filter}                    &$\surd$     &$\times$     & -0.27  & 1.92\\
HRank \cite{lin2020hrank}              &$\surd$     &$\times$     & 0.09           & 2.00     &HRank \cite{lin2020hrank}  &$\surd$     &$\times$     & 0.53  & 2.15 \\         
AMC \cite{he2018amc}                   &$\surd$     &$\times$     & 0.90           & 2.00     &FPSL\cite{mondal2023feature}              &$\surd$     &$\times$     & -0.07  & 2.16\\        
Random Pruning \cite{li2022revisiting} &$\surd$     &$\times$     & 0.94           & 2.04     & \textbf{PIC-P}   &$\surd$     &$\times$     & \textbf{-0.45} & \textbf{2.17} \\        
LRMF \cite{zhang2021filter}            &$\surd$     &$\times$     & 0.34           & 2.11     &CCM-LRR\cite{wang2023filter} &$\surd$     &$\times$     & 0.06  & 2.83\\        
FPGM \cite{he2019filter}               &$\surd$     &$\times$     & 0.33           & 2.11             &HRank \cite{lin2020hrank}                       &$\surd$     &$\times$     & 1.62  & 2.88 \\
LFPC \cite{he2020learning}             &$\surd$     &$\times$     & 0.35           & 2.12             & \textbf{PIC-P}                                 &$\surd$     &$\times$     & \textbf{0.14}  & \textbf{2.88} \\
MaskSparsity \cite{jiang2023pruning}   &$\surd$     &$\times$     & 0.31           & 2.22             &GM—3AS \cite{wang2022novel}                     &$\surd$     &$\times$     & 0.90  & 4.12 \\
\textbf{PIC-P}                         &$\surd$     &$\times$     & \textbf{-0.03} & \textbf{2.13}    &l2—3AS \cite{wang2022novel}                     &$\surd$     &$\times$     & 1.49  & 4.12 \\
FPSL\cite{mondal2023feature} &$\surd$     &$\times$     & 0.15  & 2.15                                                        &l1—3AS \cite{wang2022novel}                     &$\surd$     &$\times$     & 1.31  & 4.12 \\
HRank \cite{lin2020hrank}              &$\surd$     &$\times$     & 2.54  & 3.86                      &HBFP\cite{basha2024deep} &$\surd$     &$\times$     & 1.42  & 4.18\\
\textbf{PIC-P}                         &$\surd$     &$\times$     & \textbf{2.47}  & \textbf{4.00}    &HRank \cite{lin2020hrank}                       &$\surd$     &$\times$     & 2.73  & 4.26  \\
DNAS (Accurate) \cite{wu2018mixed}     &$\times$    &$\surd$      & -0.15  & 14.60                    & \textbf{PIC-P}                                 &$\surd$     &$\times$     & \textbf{0.69}  & \textbf{4.26}  \\
DNAS (Efficient) \cite{wu2018mixed}    &$\times$    &$\surd$      & 0.30  & 18.93                     &FSM \cite{duan2022network}                      &$\surd$     &$\times$     & 1.10  & 5.26 \\
\textbf{PIC-PQ}                        &$\surd$     &$\surd$      & \textbf{-0.63} & \textbf{33.50}   & \textbf{PIC-PQ}                                &$\surd$     &$\surd$      & \textbf{0.11}  & \textbf{43.06}   \\
PIC-P+FB                               &$\surd$     &$\surd$      & 0.75 & 54.01                      &PIC-P+FB                                        &$\surd$     &$\surd$      &  1.14 & 71.92\\
\textbf{PIC-PQ}                        &$\surd$     &$\surd$      & \textbf{0.10}  & \textbf{54.96}   & \textbf{PIC-PQ}                                &$\surd$     &$\surd$      & \textbf{0.75}  & \textbf{76.43} \\ 
\bottomrule
\end{tabular}}
\end{center}
\end{table*}

\subsection{Experiment Setting}\label{s4.1}

\subsubsection{Implementation Details} We evaluate the proposed method on the three benchmarks, CIFAR10, CIFAR100 \cite{krizhevsky2009learning} and  ImageNet \cite{deng2009imagenet}, and use standard training/test data splitting and data preprocessing on all datasets. For CIFAR10, we use the well-known ResNet56 \cite{he2016deep} and VGG16 \cite{simonyan2014very}. For CIFAR100, we use ResNet56 and MobilenetV2 \cite{sandler2018mobilenetv2}. And for ImageNet, we use ResNet18 and MobilenetV2. Of course, our method can be easily extended to other architectures. 

All models are implemented based on PyTorch. For all benchmarks and architectures, we randomly sample 6 batches of images to estimate the average rank of each feature map. The mutation rate $u = 10\%$ and the hyperparameters of the regularized evolutionary algorithm are set according to prior art \cite{chin2020towards}. We set the hyperparameter $\boldsymbol{\tau = 200}$ for all experiments. When fine-tuning, the scheme follows \cite{he2018soft} to use stochastic gradient descent \cite{nesterov1983method} for all networks with a weight decay of 5e-4. 

We use a single NVIDIA 2080Ti GPU for CIFAR10/100, and a single A100 GPU for ImageNet. For CIFAR10/100, we choose a batch size of 128 and an initial learning rate of 0.01 reduced by 5× at 30\%, 60\%, and 80\% of the epochs respectively. For ImageNet, batch size is 256 and an initial learning rate is 0.001 who reduces by 10× at 30\%, 60\%, and 80\% of the epochs respectively. We restrict selection space of available maximum bitwidth to $\left \{ 8,7,6,5,4 \right \} $ bits and the weights and activations of a layer share the same bitwidth. The value of the penalty factor $p$ usually ranges from 0 to 0.2. Based on the experience of Chakraborty \textit{et al.} \cite{chakraborty2020constructing}, we fix the first or/and last layers to full-precision in mixed precision quantization training. In addition, all three residual connections in ResNet18 require larger bits than their corresponding regular branches.

Following the experience of \cite{martinez2020training,liu2020reactnet}, the fine-tuning stage uses a distribution strategy. In the first step, we only fine-tunes a pruned model. In the second step, a network with mixed precision activation quantization and real-valued weights is fine-tuned. In the third step, the weights from the first step are inherited and the network is fine-tuned using mixed precision quantization for both weights and activations. The first step spends 400, 160, 60, 200, 100 and 100 epochs for the ResNet56 on CIFAR10, VGG16 on CIFAR10, ResNet56 on CIFAR100, MobilenetV2 on CIFAR100 networks, ResNet18 on ImageNet, MobilenetV2 on ImageNet respectively. The second step and third step both take 400 epochs for all networks on CIFAR10/100 and 100 epochs for all networks on ImageNet.

\subsubsection{Evaluation Metrics} In our experiments, classification performance is evaluated by overall accuracy, while FLOPs or BOPs compression ratio are employed to evaluate computational complexity. The BOPs metric has been used by DJPQ \cite{wang2020differentiable}. The BOPs for a convolution layer converting $C_i$ channels into $C_o$ channels is defined as:
\vskip -0.15in
\begin{gather}
\begin{aligned}
BOPs = C_iC_oK^2H_oW_oB_{W}B_{A},\label{e20}
\end{aligned}
\end{gather}
where $K$ is the kernel size, $H_o$ and $W_o$ define the output spatial size, $B_{W}$ and $B_{A}$ denote weight and activation bitwidth of current layer. The BOPs compression ratio is defined as the ratio between the total BOPs of the uncompressed and compressed models.

Without loss of generality, we use the FLOPs compression ratio to measure only the pruning effect and use the BOP compression ratio to measure the overall effect from pruning and quantization. As seen from Eq.(\ref{e20}), BOPs count is a function of both number of channels remaining after pruning and quantization bitwidth, hence BOPs compression ratio is a suitable metric to measure a CNN’s overall compression.

In all tables that follow,``P”, ``Q”  indicate ``pruning”, ``quantization”, ``Comp. ratio” indicates ``BOPs compression ratio”, and ``PIC-P+FB” indicates ``PIC-P+Fixed bitwidth”. 

\subsection{Experiments and Comparisons} \label{s4.2}

\begin{table}[h]
    \centering
    \caption{Compressing Results of VGG16 on CIFAR10 under Different Compression Ratios}
    \label{table-VGG16}
    \scalebox{1.15}{
    \begin{tabular}{ccccc}
    \toprule
    \textbf{P} & \textbf{Q}  & \textbf{Top-1(\%)} & \textbf{Top-1(\%)↓} & \textbf{Comp. Ratio} \\
    \midrule
    $\times$ & $\times$   & 93.75±0.01  & --     & --       \\ 
    $\surd$  & $\times$   & 96.06       & -2.31  & 1.43       \\
    $\surd$  & $\times$   & 96.44       & -2.69  & 1.72       \\
    $\surd$  & $\times$   & 94.20       & -0.45  & 2.17       \\
    $\surd$  & $\times$   & 93.61       & 0.14   & 2.88       \\
    $\surd$  & $\times$   & 93.06       & 0.69   & 4.26       \\  
    $\surd$  & $\surd$    & 93.97       & -0.22  & 31.58      \\
    $\surd$  & $\surd$    & 93.70       & 0.05   & 37.64      \\
    $\surd$  & $\surd$    & 93.64       & 0.11   & 43.06      \\
    $\surd$  & $\surd$    & 93.00       & 0.75   & 76.43     \\
    $\surd$  & $\surd$    & 92.00       & 1.75   & 101.71     \\ 
    \bottomrule
    \end{tabular}}
\end{table}

\begin{table*}[hb]
\caption{Comparison of Compressing ResNet18 and MobileNetV2 on ImageNet}
\label{table-rs18andmbv2}
\begin{center}
\scalebox{1.15}{
\begin{tabular}{cccccc|ccc}
\toprule
\multirow{2}{*}{\textbf{Method}} & \multirow{2}{*}{\textbf{P}} & \multirow{2}{*}{\textbf{Q}} & \multicolumn{3}{c|}{\textbf{ResNet18}} & \multicolumn{3}{c}{\textbf{MobileNetV2}} \\ \cline{4-9}
                   &       &             & \textbf{Top-1(\%)}     & \textbf{Top-1(\%)↓}     & \textbf{Comp. Ratio}          & \textbf{Top-1(\%)}    & \textbf{Top-1(\%)↓}     & \textbf{Comp. Ratio}         \\
\midrule
Full-precision  & $\times$          & $\times$          & 69.75(±0.01)       & --     & --    & 71.81(±0.09)       & --     & --      \\
FPSL\cite{mondal2023feature}          & $\surd$          & $\times$          & --              & --                 & --          & --              & 0.52                & 1.24      \\
\textbf{PIC-P}   & \textbf{$\surd$} & \textbf{$\times$}   & \textbf{69.76}     & \textbf{0.00}        & \textbf{1.33}  & \textbf{69.98}     & \textbf{1.83}        & \textbf{1.43}\\
FPSL\cite{mondal2023feature}          & $\surd$          & $\times$          & --              & --                 & --          & --              & 3.77                & 1.85      \\
SOSP\cite{nonnenmacher2022sosp}          & $\surd$          & $\times$          & --              & 0.70                 & 1.50          & --              & --                & --      \\
ACP\cite{chang2022automatic}          & $\surd$          & $\times$          & --              & 2.20                 & 1.52          & --              & --                & --      \\
EPruner \cite{lin2021network}           & $\surd$          & $\times$          & 67.31              & 2.44                 & 1.78          & --              & --                & --      \\
MFP \cite{he2022filter}           & $\surd$          & $\times$          & 67.11              & 2.64                 & 2.07          & --              & --                & --      \\
ManiDP \cite{tang2021manifold}           & $\surd$          & $\times$          & --              & --                 & --          & 69.62              & 2.19                & 2.05      \\
TWN\cite{liu2023ternary}             & $\times$          & $\surd$          & 61.80              & 7.95                 & 16.00           & --              & --                 & --       \\
SR+DR \cite{2018Ristretto}           & $\times$          & $\surd$          & 68.17              & 1.58                 & 16.00          & 61.30              & 10.51                & 16.00      \\
UNIQ \cite{Baskin_2019}            & $\times$          & $\surd$          & 67.02              & 2.73                 & 32.00           & 66.00              & 5.81                 & 32.00     \\
DFQ \cite{2019Data}             & $\times$          & $\surd$          & 65.80              & 3.95                 & 64.00           & 70.43              & 1.38                 & 16.00     \\
RQ \cite{2019Relaxed}              & $\times$          & $\surd$          & 61.52              & 8.23                 & 64.00           & 68.02              & 3.79                 & 28.44     \\
DJPQ \cite{wang2020differentiable}             & $\surd$          & $\surd$          & 69.12              & 0.63                 & 52.28           & 69.30              & 2.51                 & 43.00     \\
\textbf{PIC-Q}   & \textbf{$\times$} & \textbf{$\surd$}   & \textbf{69.20}     & \textbf{0.55}        & \textbf{47.72}  & --              & --                 & --\\
\textbf{PIC-PQ}   & \textbf{$\surd$} & \textbf{$\surd$} & \textbf{69.14}     & \textbf{0.61}        & \textbf{53.24}  & \textbf{69.40}     & \textbf{2.41}        & \textbf{43.00}\\    
\bottomrule
\end{tabular}}
\end{center}
\end{table*}

\subsubsection{Results on CIFAR10} The baselines for ResNet56 and VGG16 are 93.92\% (±0.11\%) and 93.75\% (±0.10\%) respectively, derived from multiple experiments. In Table \ref{table-CIFAR10}, for ResNet56, PIC-P outperforms comparison methods in terms of speed-up ratio and the accuracy drop when compressing 1.43×, 2.13× and 4×. \textcolor[rgb]{0,0,0}{HRank\cite{lin2020hrank}, FPGM\cite{he2019filter}, LRMF\cite{zhang2021filter}, REAF\cite{zhang2023reaf}, HBFP\cite{basha2024deep} and FPSL\cite{mondal2023feature} are hierarchically structured pruning methods based on filter importance learning, while CCM-LRR\cite{wang2023filter} and MaskSparsity\cite{jiang2023pruning} are both sparsity-training-based methods. The advantage of our method is that it can automatically determine the pruning objects through evolutionary algorithms and resource constraints without manually assigning the pruning rate for each layer, which is well guaranteed to ensure the global nature. }From the experimental results, our method can also have comparable accuracy performance to SOTAs at similar compression ratios. PIC-PQ achieves a compression of 33.50× and an accuracy improvement of 0.63\%, which is better than DNAS \cite{wu2018mixed}. 

For VGG16, compared with SOTAs at three different compression ratios, PIC-P is able to perform better in terms of accuracy drop while maintaining the same compression ratios. \textcolor[rgb]{0,0,0}{FSM\cite{duan2022network} is also a hierarchically structured pruning method based on filter importance learning. Similarly, our approach performs similarly on the VGG16 network, with a good trade-off between accuracy preservation and compression ratio. Even at a compression ratio of 2.17, there is a 0.45\% improvement in accuracy after compression, outperforming the CCM-LRR\cite{wang2023filter} and FPSL\cite{mondal2023feature} on all fronts. When the compression is relatively large, the accuracy of our method decreases by 0.69\% when compressed by 4.26×, while the latest method HBFP\cite{basha2024deep}’s accuracy decreases by 1.10\% when compressed by 4.18×. Here our method is superior.}

More notably, we also evaluate PIC-P in combination with fixed bitwidth quantization as a two-stage approach and make sure that the pruning rate is very close. In comparison, PIC-PQ is able to maintain better accuracy at a similar compression ratio when quantization bitwidth searching is performed, which demonstrates the effectiveness of our search-based approach. 

Besides, we also give the compression results of VGG16 on CIFAR10 under different BOPs compression ratios. We want to understand how our design affects the performance of the classification network. Specifically, we try PIC-P as well as PIC-PQ for VGG16 on CIFAR10 and configure five different compression ratios for PIC-P and PIC-PQ respectively, which users can use as a reference to choose according to their actual needs. The specific results are shown in Table \ref{table-VGG16}. We can see that with only pruning when compressing from 1.43× to 2.17×, the accuracy of each network improves, which indicates that the network does have a degree of redundancy, while we remove it in an interpretable way. In addition, joint learning method can compress better than pruning alone within a range of compression ratios.

\subsubsection{Results on ImageNet} Table \ref{table-rs18andmbv2} provides a comparison of PIC-PQ with other methods for compressing ResNet18 and MobileNetV2 on ImageNet. The results show a smooth trade-off achieved by our method between accuracy and BOPs reduction. \textcolor[rgb]{0,0,0}{For a similar range of compression ratios on ResNet18, PIC-P has better accuracy retention compared to the two comparison methods. SOSP\cite{nonnenmacher2022sosp} devises two novel saliency-based methods for second-order structured pruning which include correlations among all structures and layers. ACP\cite{chang2022automatic} is an automatic channel pruning method via clustering and swarm intelligence optimization. Compared to these two approaches, our advantage lies not only in combining quantization, but also in considering the global ranking of each filter. }Compared with some fixed-bit quantization schemes, PIC-PQ achieves a significantly larger BOPs reduction. 

MobileNetV2 has been shown to be very sensitive to quantization \cite{2019Data}. \textcolor[rgb]{0,0,0}{Despite this, PIC-P and PIC-PQ is able to balance the compression ratio and accuracy well during compression, and performs more consistently than FPSL\cite{mondal2023feature}.} PIC-PQ is able to compress MobileNetV2 with a large BOPs compression ratio. PIC-PQ achieves a 43× BOPs reduction within 2.41\% accuracy drop over DJPQ, where we set a similar global pruning rate.

\begin{table}[h]
\caption{Pruning Results of ResNet56 on CIFAR100}
\vskip -0.2in
\label{table-RS56-P}
\begin{center}
\scalebox{1.0}{
\begin{tabular}{c|cccc}
\toprule
\textbf{Method}     & \textbf{Baseline(\%)} & \textbf{Pruned(\%)}     & \textbf{Top-1(\%)↓} & \textbf{FLOPs↓(\%)} \\
\midrule
REAF\cite{zhang2023reaf}      &69.58     &70.04      & -0.46      & 38.0 \\
REAF\cite{zhang2023reaf}      &69.58     &69.53      & 0.05      & 50.0 \\
AMC\cite{he2018amc}       &70.66     &68.92      & 1.74      & 50.0 \\
LFPC\cite{he2020learning}        &--     &--      & 0.58      & 51.6 \\
SFP\cite{he2018soft}        &--     &--      & 2.61      & 52.6 \\
FPGM\cite{he2019filter}        &--     &--      & 1.75      & 52.6 \\
LRMF\cite{zhang2021filter}       &69.67     &69.23      & 0.44      & 52.6 \\
\textbf{PIC-P}     &\textbf{70.66}     &\textbf{71.10}      & \textbf{-0.44} & \textbf{54.0} \\ 
\bottomrule
\end{tabular}}
\end{center}
\vskip -0.15in
\end{table}

\subsubsection{Results on CIFAR100} \textcolor[rgb]{0,0,0}{Table \ref{table-RS56-P} demonstrates the comparison of the results of different methods for pruning the ResNet56 network on CIFAR100. The results in Table \ref{table-RS56-P} show that PIC-P significantly outperforms the classical filter pruning algorithms AMC\cite{he2018amc}, LFPC\cite{he2020learning}, SFP\cite{he2018soft}, FPGM\cite{he2019filter} and LRMF\cite{zhang2021filter} when FLOPs are reduced by about 50.0\%. The latest REAF\cite{zhang2023reaf} shows a small decrease in pruning accuracy when the FLOPs drop by about 50.0\% and 52.6\%, respectively, while PIC-P shows a small increase in pruning accuracy when the FLOPs drop by 54.0\%. In addition, compared with REAF\cite{zhang2023reaf}, PIC-P achieves greater compression performance when the accuracy after pruning increases by a similar value. }

\begin{table}[h]
\caption{Compressing Results of MobilenetV2 on CIFAR100}
\label{table-MBV2}
\begin{center}
\scalebox{1.0}{
\begin{tabular}{c|ccccccc}
\toprule
\textbf{Method}     & \textbf{P} & \textbf{Q}  & \textbf{Top-1(\%)}     & \textbf{Top-1(\%)↓} & \textbf{Comp. Ratio} \\
\midrule
 Full-precision      &$\times$     &$\times$      & 75.12      & --                & --   \\
 PIC-P+FB            &$\surd$     &$\surd$      & 65.41          & 9.71                & 54.44   \\
\textbf{PIC-PQ}     &$\surd$     &$\surd$      & \textbf{72.85} & \textbf{2.27}       & \textbf{55.18} \\
\bottomrule
\end{tabular}}
\end{center}
\vskip -0.15in
\end{table}

As we can see from Table \ref{table-MBV2}, for MobilenetV2, PIC-PQ performs better than PIC-P+FB (Weight/Activation bitwidth (W/A) is 8/6) due to learning a better quantization strategy and preserving a higher quantization accuracy for some initial key layers of MobilenetV2.

\begin{table}[h]
\caption{Compressing Results of ResNet56 on CIFAR100.}
\label{table-RS56}
\begin{center}
\scalebox{1.0}{
\begin{tabular}{c|ccccccc}
\toprule
\textbf{Method}     & \textbf{P} & \textbf{Q}  & \textbf{Top-1(\%)}     & \textbf{Top-1(\%)↓} & \textbf{Comp. Ratio} \\
\midrule
Full-precision      &$\times$     &$\times$      & 70.66      & --                & --   \\
PIC-P+FB            &$\surd$     &$\surd$      & 66.82          & 3.84                & 40.61   \\
\textbf{PIC-PQ}     &$\surd$     &$\surd$      & \textbf{68.92} & \textbf{1.74}       & \textbf{42.68}   \\ 
\bottomrule
\end{tabular}}
\end{center}
\vskip -0.15in
\end{table}

Similarly, we also compare PIC-PQ with PIC-P+FB on ResNet56 in Table \ref{table-RS56}. PIC-PQ continues to have a lower accuracy loss than PIC-P+FB (W/A is 8/8) on ResNet56.

\section{Discussion}\label{s5}

\subsection{Validity Analysis of the Physics Inspired Criterion (PIC)} \label{s5.1}

\begin{table}[h]
\centering
\caption{The Effect of PIC for Compressing ResNet56 on CIFAR10}
\label{table-Ablation1}
\scalebox{1.1}{
\begin{tabular}{c|ccc}
\toprule
\textbf{PIC} & \textbf{Comp. Ratio} & \textbf{Top-1(\%)} & \textbf{Top-1(\%)↓ }\\
\midrule
$\surd$       & 1.43 & 94.01 →94.32 & -0.31  \\
$\times$       & 1.43 & 94.03 →93.36 & 0.67  \\ 
$\surd$       & 2.13 & 94.03 →94.06 & -0.03  \\
$\times$ & 2.13 & 94.03 →92.02 & 2.01 \\ 
$\surd$       & 4.00 & 94.03 →91.56 & 2.47 \\
$\times$ & 3.33 & 94.03 →89.68 & 4.35 \\ 
\bottomrule
\end{tabular}}
\end{table}

We investigate the effect of the PIC. We first turn off PIC-Q and then perform experiments with and without (default $\boldsymbol{a=1}$,$\boldsymbol{b=0}$) the PIC. The results are presented in Table \ref{table-Ablation1}, showing that in the absence of PIC for learning the optimal $\boldsymbol{a-b}$ pairs, the global ranking of the filters will be unguided and the accuracy decreases significantly at similar compression ratios compared to adopting PIC.

\subsection{Influence of the Fine-tuning Iterations in Searching the Optimal a-b Pairs} \label{s5.2}

\begin{figure}[htbp]
\centering
\includegraphics[scale = 0.3]{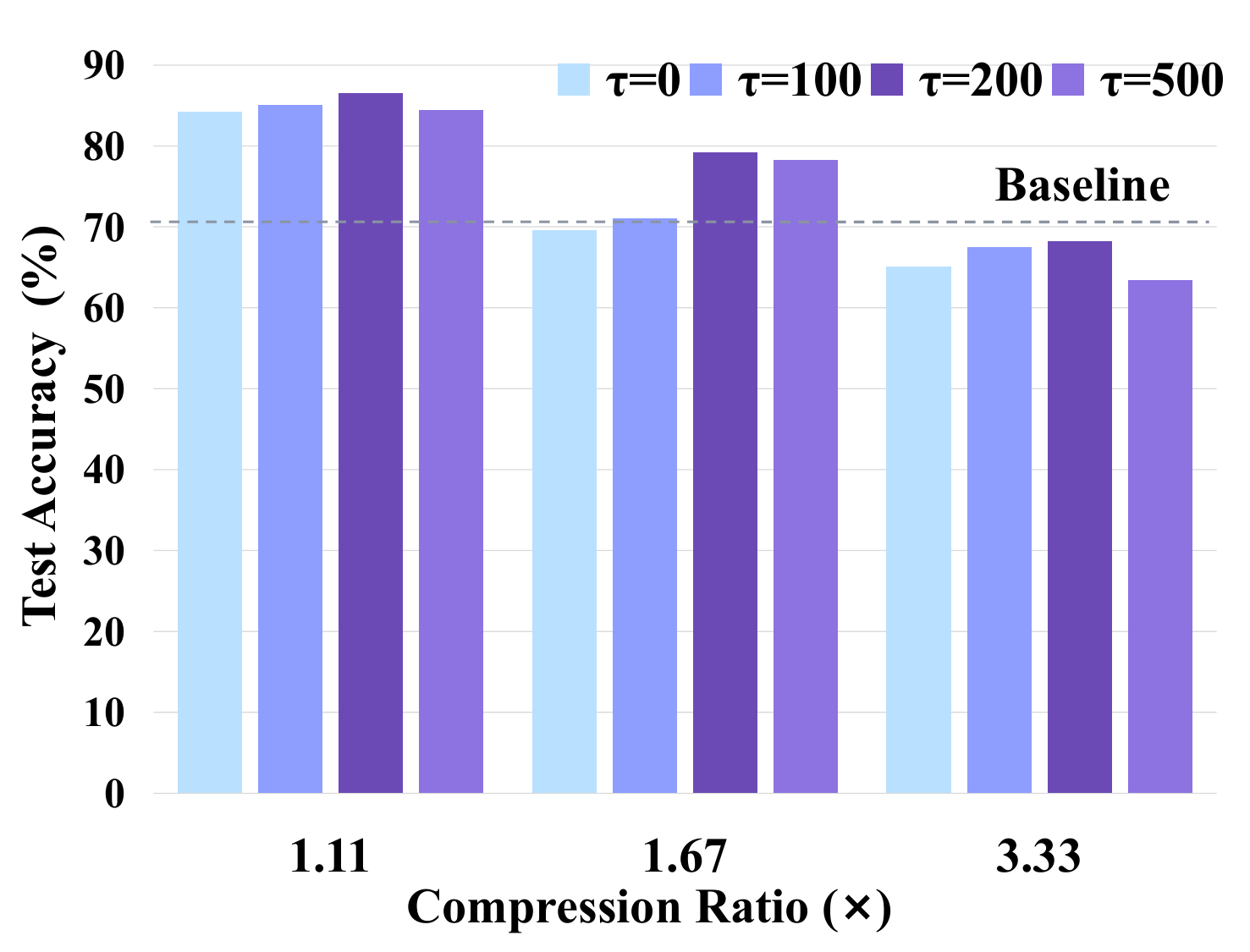}
\caption{Influence of the $\boldsymbol{\tau}$ when compressing ResNet56 on CIFAR100.}
\label{p5}
\end{figure}

The actual fine-tuning process is simulated in searching the optimal $\boldsymbol{a-b}$ pairs, where we further analyze the impact of the steps $\boldsymbol{\tau}$. We choose different $\boldsymbol{\tau}$ at three similar compression ratios, and once the optimal $\boldsymbol{a-b}$ pairs are learned, we compress ResNet56 on CIFAR100, where still only PIC-P is used. In the previous, we mention that $\boldsymbol{\tau}$ in searching the optimal $\boldsymbol{a-b}$ pairs approximates the fully fine-tuning steps $\boldsymbol{\hat{\tau}}$. We expect that the closer  $\boldsymbol{\tau}$ to $\boldsymbol{\hat{\tau}}$, the better the optimization of the $\boldsymbol{a-b}$ pairs. In this experiment, 60 epochs are iterated, so that $\boldsymbol{\hat{\tau}}$ is fixed at 21120 gradient steps. As shown in Figure \ref{p5}, $\boldsymbol{\tau}=200$ works well. The results are consistent with our intuition that continuing to increase  $\boldsymbol{\tau}$ as we approach $\boldsymbol{\hat{\tau}}$ produces a decreasing effect.

\subsection{Search a-b Pairs Every Time or Once} \label{s5.3}

\begin{table}[htbp]
\centering
\caption{Comparison of Compressing VGG16 on CIFAR10 under \textbf{``Just-searching-once”} and \textbf{``Searching-every-time”}}
\label{table-Ablation2}
\scalebox{1.0}{
\begin{tabular}{cc|cc|cc}
\hline
\multirow{2}{*}{\textbf{P}} & \multirow{2}{*}{\textbf{Q}} & \multicolumn{2}{c|}{\textbf{searching-every-time}} & \multicolumn{2}{c}{\textbf{just-searching-once}} \\ \cline{3-6}
                   &                    & \textbf{Top-1(\%)}          & \textbf{Comp. Ratio}          & \textbf{Top-1(\%)}         & \textbf{Comp. Ratio}         \\ \hline
$\surd$                & $\times$                & 96.04              & 1.43                 & 96.06             & 1.43                \\
$\surd$                & $\times$                & 94.20              & 2.17                 & 94.18             & 2.17                \\
$\surd$                & $\times$                & 91.34              & 4.55                 & 91.32             & 4.55                \\
$\surd$                & $\surd$                & 93.95              & 31.60                & 93.97             & 31.58               \\
$\surd$                & $\surd$                & 93.64              & 43.06                & 93.58             & 44.28               \\
$\surd$                & $\surd$                & 91.38              & 119.18               & 92.00             & 101.71              \\ 
\bottomrule
\end{tabular}}
\end{table}
In Table \ref{table-Ablation2}, \textcolor[rgb]{0,0,0}{\textbf{ ``searching-every-time”} }represents searching $\boldsymbol{a-b}$ pairs separately when compressing at different compression ratios. While \textbf{``just-searching-once”} means that after searching for the optimal solution of $\boldsymbol{a-b}$ pairs at a compression ratio to get the global ranking of the filters, you only need to use the obtained ranking information at other compression ratios. Here PIC-P and PIC-PQ use the prior knowledge searched at compression ratios of 1.43× and 31.58× respectively. Through Table \ref{table-Ablation2}, we find that both routes produce similar accuracy. 

In practice, our method relies on the assumption that\textit{ the best-performing narrow network is a subset of the best-performing wide network}. There are many model compression methods that achieve a given compression ratio by manually or automatically changing the filter number in different layers, which means that there may be: compared to the best-performing large network, the best performing small network has a larger number of filters in some layers, but a smaller number of filters in some other layers. However, in our method, for a given network, it is only necessary to search and learn once, and this can be used to obtain compressed networks with different compression ratios. This assumption not only simplifies the tedious process of learning filter ranking, but also allows the user to have multiple choices of compression ratios under \textbf{``just-searching-once”}, which improves the efficiency of model compression. 

\begin{figure}[h]
\centering
\includegraphics[scale = 0.4]{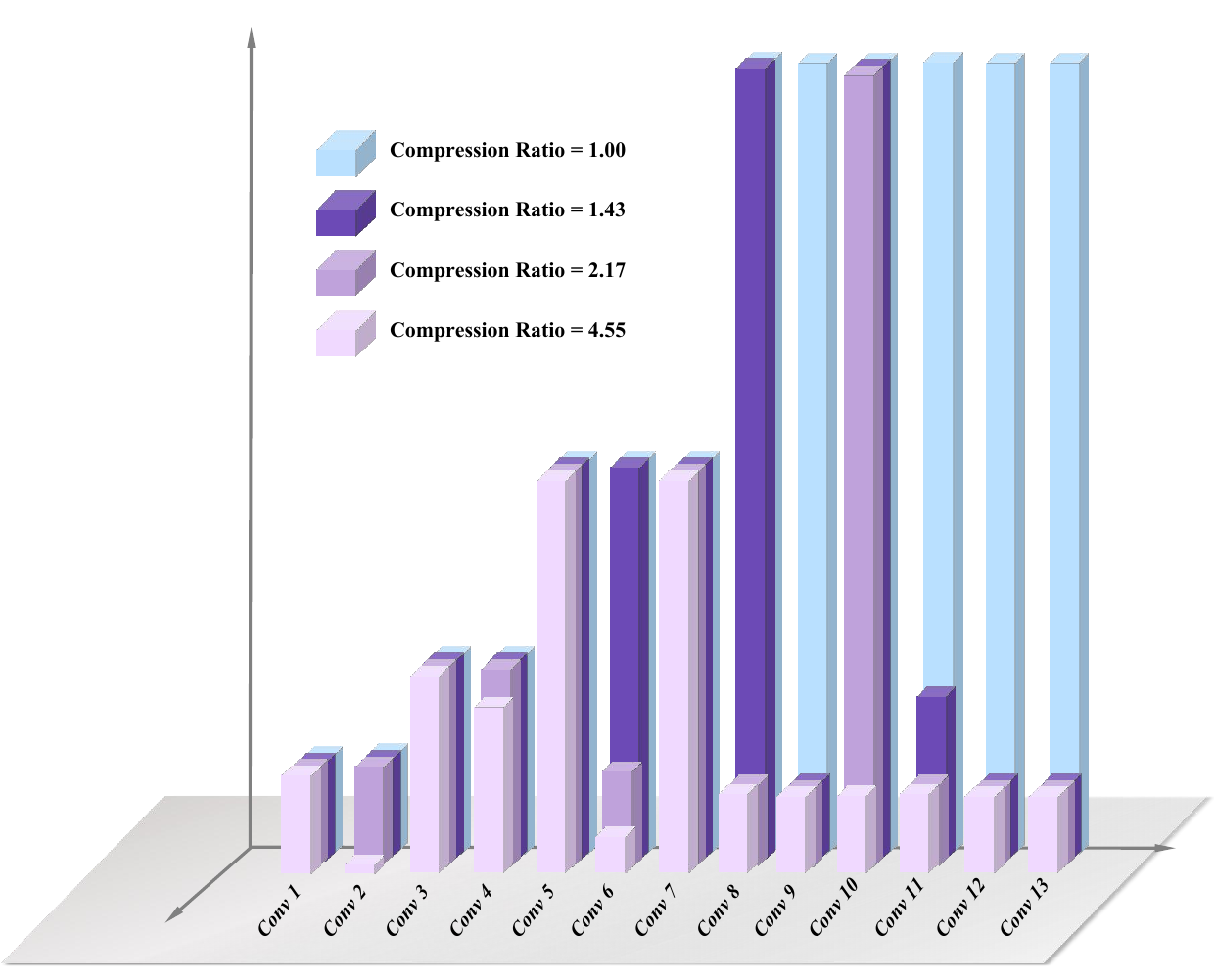}
\caption{Comparison of the filter number remaining in each layer of VGG16 under \textcolor[rgb]{0,0,0}{\textbf{``just-searching-once”}}.}
\label{p6}
\end{figure}

Figure \ref{p6} plots a comparison of the filter number remaining in each layer of the VGG network under \textbf{``just-searching-once”}. From the Figure \ref{p6}, we can not only see that our actual compression results satisfy the above subset of assumptions, but also confirm an opinion that the top layer is more stable and more inclined to be compressed than the lower layer.

\section{Conclusion}\label{s6}
In this paper, we propose a novel physics inspired criterion for pruning-quantization joint learning (PIC-PQ), where we make the first attempt to draw an analogy between ED and MC. A physics inspired criterion for ranking filters' importance globally is explored from the analogy. We are well in line with the feature interpretability. Specifically, learnable deformation scale and FP in PIC are derived from Hooke’s law in ED. Besides, we further extend PIC with a relative shift variable to rank filters globally. An objective function is also put forward additionally from a mathematical theory perspective to demonstrate the viability of PIC. To ensure feasibility and flexibility, available maximum bitwidth and penalty factor are introduced in quantization bitwidth assignment. Experiments on benchmarks show that PIC-PQ is able to achieve a good trade-off between accuracy and BOPs reduction.
\bibliographystyle{IEEEtran}  
\bibliography{sample}  

\end{document}